\pdfoutput=1
\documentclass[11pt]{article}

\usepackage[final]{acl}

\usepackage{booktabs} % For formal tables
\usepackage{tabularx} % For adjustable-width tables
\usepackage{array}    % For new column types
\usepackage{ragged2e} % Provides \justifying command

\usepackage[most]{tcolorbox}
\usepackage{times}
\usepackage{latexsym}
\usepackage{multirow}
\usepackage{inconsolata}
\usepackage{multirow}
\usepackage{graphicx}
\usepackage{color}
\usepackage{hyperref}
\usepackage{microtype}
\usepackage{lscape}
\usepackage{longtable}
\usepackage{booktabs} % For professional looking tables
\usepackage{tabularx} % For adjustable width tables
\usepackage{lipsum} % Dummy text
\usepackage{times}
\usepackage{siunitx}
\usepackage{makecell}
\usepackage{latexsym}
\usepackage{amsmath}
\usepackage{mdframed,lipsum}
\usepackage[T1]{fontenc}
\usepackage[utf8]{inputenc}

\usepackage{microtype}

\usepackage{inconsolata}

\title{Beyond Binary Gender Labels: Revealing Gender Biases in LLMs through Gender-Neutral Name Predictions}

\author{Zhiwen You$^1$\thanks{Equal Contribution.}, HaeJin Lee$^1$\footnotemark[1], Shubhanshu Mishra$^2$,\\ 
\textbf{Sullam Jeoung$^1$, Apratim Mishra$^1$, Jinseok Kim$^3$, Jana Diesner$^1, ^4$} \\
$^1$ University of Illinois Urbana-Champaign \quad 
$^2$ https://shubhanshu.com \\
$^3$ University of Michigan - Ann Arbor \quad 
$^4$ Technical University of Munich \\
$^1$ \texttt{\{zhiweny2, haejin2, sjeoung2, apratim3\}@illinois.edu} \\
$^2$ \texttt{mishra}\texttt{@shubhanshu.com} \quad $^3$ \texttt{jinseokk}\texttt{@umich.edu} \quad
$^4$ \texttt{jana.diesner}\texttt{@tum.de}}

\begin{document}
\maketitle
\begin{abstract}
% Numerous studies have demonstrated the presence of bias in gender prediction tasks. These biases can significantly influence a range of downstream applications---from personalized recommendations to investigating the presence of gender biases in academia using bibliometrics. In this paper, we expand the scope of gender labels by introducing "neutral" as an additional category and examine the performance of various Large Language Models (LLMs) in gender prediction given first names.  These applications, ranging from personalized recommendations to investigating gender biases in academia through bibliometrics, could benefit from more accurate and inclusive gender classification. 
% + use word: bias in model prediction in gender-neutral names

Name-based gender prediction has traditionally categorized individuals as either female or male based on their names, using a binary classification system. That binary approach can be problematic in the cases of gender-neutral names that do not align with any one gender, among other reasons. Relying solely on binary gender categories without recognizing gender-neutral names can reduce the inclusiveness of gender prediction tasks. We introduce an additional gender category, i.e., ``neutral'', to study and address potential gender biases in Large Language Models (LLMs). We evaluate the performance of several foundational and large language models in predicting gender based on first names only. Additionally, we investigate the impact of adding birth years to enhance the accuracy of gender prediction, accounting for shifting associations between names and genders over time. Our findings indicate that most LLMs identify male and female names with high accuracy (over 80\%) but struggle with gender-neutral names (under 40\%), and the accuracy of gender prediction is higher for English-based first names than non-English names. The experimental results show that incorporating the birth year does not improve the overall accuracy of gender prediction, especially for names with evolving gender associations. We recommend using caution when applying LLMs for gender identification in downstream tasks, particularly when dealing with non-binary gender labels\footnote{Our code is available at \url{https://github.com/zhiwenyou103/Beyond-Binary-Gender-Labels}.}.

\end{abstract}

% Gender classifiers are applied in various downstream tasks, e.g., distinguishing author's express sentiment by analyzing their storytelling behavior to classify different genders based on their names \cite{jentzsch2022gender}.
\section{Introduction} 
% Gender prediction serves as a foundation for numerous downstream applications, including investigating gender bias, personalized marketing, content recommendation, targeted advertising, gender-based sentiment analysis, and social network analysis \cite{diesner2009he, ross2022women, jentzsch2022gender, teich2022citation, liu2023gender}. This task involves using computational solutions (e.g., Genderize.io\footnote{\url{https://genderize.io/}}, Namsor\footnote{\url{https://namsor.app/}}, Gender API\footnote{\url{https://gender-api.com/}}, or machine learning (ML) models) to estimate a name's gender based on one or more pieces of information, such as the constituents of a name and demographics. Gender disparities in academia have drawn significant attention, prompting researchers to investigate gender gaps in professorship roles and authorship \cite{vanhelene2024inferring}. 

% Added today
\begin{figure}
  \centering
  \includegraphics[scale=0.48]{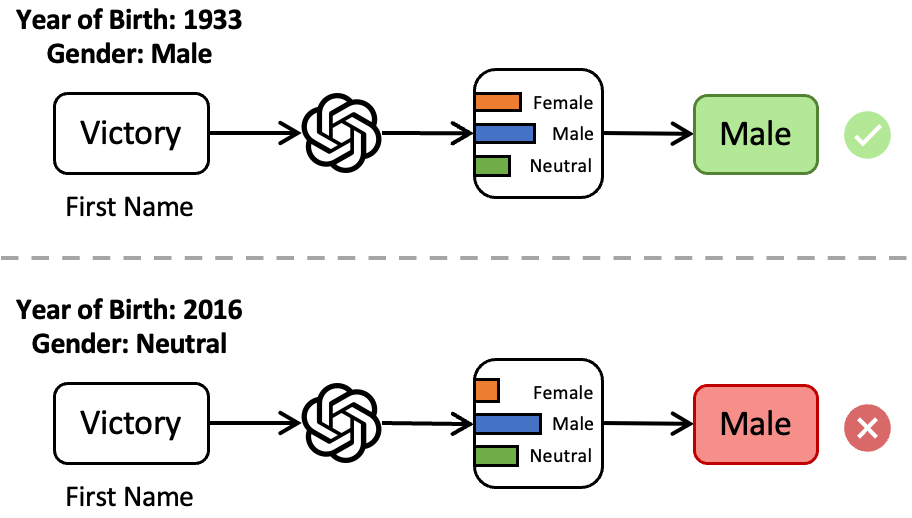}
  \caption{Example of an LLM predicting different gender labels over time for the same first name. ``Victory'' was labeled Male in 1933, and the LLM predicted it correctly. However, by 2016, the name had become predominantly gender-neutral, but the LLM still incorrectly predicted it as Male.}
  \label{diagram}
\end{figure}

Name-based gender prediction is the task of identifying the most likely gender label for a given name. This task, while not reflective of the true gender identify of the individual, is often useful for aggregate downstream analysis and as a demographic feature for predictive models. Prior work has utilized name-based gender prediction to investigate gender bias in scientific productivity, citation practices, information extraction systems, personalized marketing, content recommendation, targeted advertising, gender-based sentiment analysis, and social network analysis \cite{diesner2009he, ross2022women, jentzsch2022gender, teich2022citation, liu2023gender,lariviere_bibliometrics_2013,mishra2020assessingdemographicbiasnamed,mishra2018self,vanhelene2024inferring}. Most prior work has utilized computational tools (e.g., Genderize.io\footnote{\url{https://genderize.io/}}, Namsor\footnote{\url{https://namsor.app/}}, Gender API\footnote{\url{https://gender-api.com/}}, or machine learning (ML) models) or datasets (e.g., US SSN) to assign probabilities of a name (along with other features like demographics, time) likely to be a male or a female. Since name-based gender is used both as a feature in downstream systems and an indicator of demographic representation, it can lead to both measurement bias and representational bias as identified in the framework proposed by \citet{Harini2021}.

A prevalent challenge in contexts utilizing inferred gender is the practice of treating gender as a binary construct, strictly categorizing names as either male or female \cite{chatterjee2021gender, pilkina2022gender}. This reliance on binary labels likely stems from historical and societal norms that often only recognize these two categories. Binary representations can reinforce existing gender biases and exclude non-binary and gender-diverse individuals, hindering their representation and understanding \cite{krstovski2023inferring, dinh2023we,mishra2018self} in algorithm design and data annotation. The presence of gender-neutral names, as defined by \citet{barry2014unisex}, further complicates this issue. These names, frequently assigned to both genders, contradict the binary classification system, leading to potential inaccuracies and misrepresentations in data and processes reliant on gender predictions.

This study aims to answer the following research questions to examine one of many aspects of gender biases in LLMs concerning gender prediction, especially for gender-neutral names and gender labels that change over time (Figure~\ref{diagram}). 

\textbf{RQ1.} How does the performance of autoregressive LLMs versus fine-tuned foundation language models compare when predicting gender categories (i.e., female, male, and neutral) given first names? 
% RQ2: Considering that baby names may recur across different years, can including the birth year enhance the accuracy of LLMs in gender prediction based on first names? 

\textbf{RQ2.} How does adding the birth year impact gender prediction accuracy?

\textbf{NOTE:} In the context of this research, we are only interested in studying the likelihood of a name being identified as \texttt{Male}, \texttt{Female}, and \texttt{Neutral}. As highlighted in \citet{ImageCrop2021}, predictive models cannot be accurate about demographic attributes, and it is best to rely on individual responses to assign sensitive demographic attributes e.g. gender, however, they can be useful at the aggregate level, which is the focus of this work.

% \textbf{RQ3.} What are the temporal trends in gender predictions made by LLMs across three countries? (Unisex)
% How question. 
% How does providing birth year as an additional info 
% Additionally, is there a bias in the prediction of unisex names by LLMs, possibly due to a lack of long-term training data that includes ``unisex” as a gender label? 

\section{Related Work} 
In the gender prediction task, models are trained to predict or classify gender labels based on various input features, such as first or last names, country information, behavioral data, or textual content from social media activity \cite{liu2013s, tang2011s, to2020gender}. Consequently, the accuracy of gender prediction can impact the validity of research findings and derived implication, such as policies. In other words, inaccurate gender prediction can distort results and lead to misunderstandings of gender-related biases. Moreover, the reliance on binary gender categorizations constrains the nuanced understanding of bias and the representation of individuals. Therefore, ensuring accurate and unbiased gender prediction is essential as it can impact the fairness and effectiveness of downstream applications. 

Previous studies found prevalent biases in NLP-based gender prediction using gender-predicting software tools \cite{misa2022gender, alexopoulos2023gender}, which failed to appropriately capture the fact that gender exists on a non-binary scale. While most studies of bias in gender prediction relied on binary gender labels \cite{teich2022citation, liu2023gender}, some studies have gone beyond binary labels by introducing an additional category for names that were not strictly associated with either female or male genders \cite{lariviere2013bibliometrics, mishra2018self, pinheiro2022women}. For instance, \citet{krstovski2023inferring} categorized names that appeared as both female and male as ``gender ambiguous''. Additionally, most prior work on gender prediction used names as the only input feature \cite{jia2019gender, hu2021s, pham2023gendec}, while others such as \citet{blevins2015jane} and \citet{misa2022gender} inferred the gender of first names using historical datasets with multilpe features.

% Although \citet{krstovski2023inferring} involved names that appeared as both female and male were categorized as ``gender ambiguous'', these biases predominantly focused on binary gender categories: female and male, which neglected gender-neutral labels. 

Recent advances in deep learning (DL) have produced pre-trained language models like BERT \cite{devlin2018bert}, CharBERT \cite{ma-etal-2020-charbert}, and RoBERTa \cite{liu2019roberta}, which have been widely used for gender prediction. For example, \citet{hu2021s} found that using the user's name achieved higher gender prediction accuracy than using other features (e.g., website page views and clicks) in both ML and DL models, while \citet{jia2019gender} and \citet{pham2023gendec} demonstrated the effectiveness of BERT-based models for gender prediction for Japanese and Chinese names. Despite these developments, few studies focused on gender prediction using autoregressive models like ChatGPT \cite{chatgpt} and Llama 2 \cite{touvron2023llama}. The increasing application of LLMs for gender prediction \cite{kotek2023gender, rhue2024evaluating} underscores the need to evaluate the limitations of LLMs, particularly for gender-neutral names. For example, \citet{michelle2023gender} used a prompting approach with ChatGPT to predict the gender of Olympic athletes, showing ChatGPT performed at least as well as common commercial tools (i.e., Gender-API and Namsor) and often outperforms them on a binary gender scale. In this paper, we conducted experiments beyond prior approaches by introducing the gender-neutral label and using three Social Security Administration (SSA) baby name datasets to investigate gender biases by predicting non-binary gender labels.

\section{Experiments}
This section discussed the datasets, pre-processing, experimental design, and how we compared various models for name-based gender prediction. 

\subsection{Data} \label{data}
\textbf{Dataset Pre-processing.} We re-used three datasets of first names of children: one from the SSA of the US\footnote{\url{https://www.ssa.gov/oact/babynames/limits.html}}, one from the province of Alberta, Canada\footnote{\url{https://ouvert.canada.ca/data/dataset}}, and one from France\footnote{\url{https://www.insee.fr/fr/statistiques/7633685?sommaire=7635552}}. Each dataset included first names, gender (female or male), and birth year. To identify and associate the gender label for each name, we counted how often each name appeared with its associated gender labels (i.e., female or male) and year of birth for a specific year. For example, if the name ``Harry'' appeared five times as female and 15 times as male in a specific year, we calculated the gender ratios for that year as 25\% female and 75\% male. Using these ratios, we labeled the first names with the associated gender labels according to the following rule-set: if a first name was at least 10\% female and 10\% male representation in a given year, we labeled the name as neutral. For first names with at least 85\% female representation, we labeled the names as female gender label. Similarly, for the first names with at least 85\% male, we labeled the names as male. 

% For each dataset, we calculated the proportion of first names with neutral gender labels by dividing the number of gender-neutral names by the total number of first names for each year. 
% This enables us to check how the prevalence of gender-neutral names has changed over time. We found that gender-neutral names have increased in both the US and Canada SSA datasets over time. The France SSA dataset showed an increase in XXX from 1900 to 1940, followed by a decrease and another increase starting in 1980. %Figure 

Due to the scarcity of gender-neutral names in our relabeled datasets from the 1900s, we needed to balance the number of names by gender to ensure fair comparisons in our experiments. We achieved this by sampling an equal number of female, male, and neutral names each year in the relabeled datasets. Specifically, we randomly selected 300 names per gender for each year from 1914 to 2022 from the US SSA dataset. In the Canada SSA dataset, where gender-neutral names were rare before 2000 (less than five first names per year) but increased in recent years (after 2010), we sampled 273 names per gender for each year from 2013 to 2020. Similarly, the France SSA dataset had few gender-neutral names in the early 1900s. Therefore, we selected 32 names per gender for each year from 1908 to 2022. Additional details on the dataset statistics can be found in Appendix~\ref{dataset-creation}. We used these balanced datasets for all the experiments in Table~\ref{tab2:gender-performance}.

\begin{table}
\centering
% \footnotesize
% \tiny
\footnotesize
\scalebox{0.78}{
\begin{tabular}{cccc}
% \hline
\toprule[1.5pt]
\textbf{First Names} & \textbf{Gender 1 (year)} & \textbf{Gender 2 (year)} & \textbf{Gender 3 (year)}\\
\midrule
Arlie & Male (1971) & Neutral (1980) & -\\
% \midrule
Hasani & Neutral (1983) & Male (2000) & -\\
Neer & Male (2014) & Neutral (2018) & -\\
CARMEL & Neutral (1920) & Male (1951) & -\\
FIDELE & Neutral (1918) & Female (1945) & -\\
Morley & Female (2013) & Neutral (2015) & Female (2017) \\
Victory & Male (1933) & Female (2000) & Neutral (2016)\\
Carmin & Male (1924) & Neutral (1958) & Female (2021) \\
\bottomrule[1.5pt]
\end{tabular}}
\caption{\label{gender-cases}
Examples of first names that were labeled as different genders over the years.
}
\end{table}

\textbf{Dynamic gender label datasets.} 
% Each dataset contains changeable genders of first names, meaning the same name appears multiple times with different gender labels (Table~\ref{gender-cases}). In Table~\ref{gender-cases}, we provided cases of different first names labeled as various genders over the years. To further examine how the performance of gender prediction varies for genders labeled differently across birth years and analyze the temporal trend of LLMs predictions in different model settings (i.e., zero-shot and few-shot settings), we constructed three changeable gender datasets mainly based on the test sets of our created non-binary name gender datasets (see Appendix~\ref{dataset-creation} for more details). 
We observed that each balanced SSA dataset included first names labeled with different genders over the years, as shown in Table~\ref{gender-cases}. For example, \texttt{Victory} was recorded as a male name in 1933, a female name in 2000, and as a gender-neutral name in 2016 (Figure~\ref{diagram}). To further analyze the gender prediction performance of LLMs on first names with varying gender labels over time, we created a dynamic gender label dataset for each country. We selected first names with dynamic gender labels (i.e. names for which the gender association changes over time) from the test set of each balanced SSA dataset. The dynamic gender label datasets were used in the experiments of Table~\ref{tab:duplicated-performance}. The distribution of these dynamic gender labels is detailed in Appendix~\ref{dataset-creation}.

% and to analyze temporal trends of LLMs predictions in different model settings (i.e., 0-shot and 5-shot settings)

% This allow us to analyze the impact of temporal changes in naming conventions on model accuracy. 

\begin{table*}[!htb]
\centering
\footnotesize
\begin{tabular}{cccccc|cccc|c} 
\toprule[1.5pt]
 % & & \multirow{3}{*}{First Name} & Micro F1 & Acc. & Macro F1 & Micro F1 & Acc. \\
% \multicolumn{2}{c{>{\centering\arraybackslash}p{2cm}}} & First & First+Country\\
\multicolumn{1}{c}{}&\multicolumn{1}{c}{}&\multicolumn{4}{c}{First Name}&\multicolumn{4}{c}{First Name + Year}\\
\midrule
Datasets & Models & Male & Female & Neutral & Acc. & Male & Female & Neutral & Acc. & Avg. \\
\midrule
\multirow{9}{*}{US SSA} & BERT & 84.46 & \textbf{89.30} & \textbf{90.55} & \textbf{88.10} & \textbf{86.64} & \textbf{90.98} & 91.13 & \textbf{89.58} & \textbf{88.84}\\
& RoBERTa & 83.76 & 87.80 & 90.00 & 87.19 & 85.05 & 88.53 & 90.95 & 88.18 & 87.69\\
& CharRoBERTa & \textbf{84.62} & 88.81 & 88.99 & 87.47 & 83.55 & 88.59 & \textbf{91.96} & 88.03 & 87.75\\
\cmidrule{2-11}
& GPT-3.5 & 91.62 &\textbf{ 96.70} & 15.99 & 68.10 & 94.68 & \textbf{96.30} & 14.37 & \textbf{68.45} & 68.28\\
& Llama 2 & 1.93 & 6.42 & \textbf{99.66} & 36.00 & 16.48 & 36.97 & \textbf{90.37} & 47.94 & 41.97\\
& Llama 3 & \textbf{94.80} & 94.83 & 13.03 & 67.55 & 95.29& 95.26 & 6.09 & 65.55 & 66.55\\
& Mixtral-8x7B & 64.62 & 85.81 & 53.30 & 67.91 & 61.38 & 78.44 & 56.42 & 65.41 & 66.66 \\
& Claude 3 Haiku & 91.50 & 93.67 & 30.00 & \textbf{71.72} & \textbf{96.30} & 93.46 & 6.97 & 65.58 & \textbf{68.65}\\
\midrule
\multirow{9}{*}{Canada SSA} & BERT & 70.98 & 73.21 & \textbf{82.14} & \textbf{75.45} & \textbf{74.11} & 74.55 & 74.11 & \textbf{75.15} & \textbf{75.30}\\
& RoBERTa & \textbf{72.77} & 75.00 & 73.66 & 73.81 & 67.86 & 75.00 & \textbf{76.34} & 73.07 & 73.44\\
& CharRoBERTa & 71.43 & \textbf{76.34} & 71.88 & 73.21 & 69.20 & \textbf{76.34} & 74.11 & 73.21 & 73.21\\
\cmidrule{2-11}
& GPT-3.5 & 82.14 & \textbf{86.61} & 27.68 & 65.48 & \textbf{83.93} & 83.93 & 28.12 & 65.33 & 65.41\\
& Llama 2 & 1.79 & 11.16 & \textbf{100.00} & 37.65 & 0.45 & 9.82 & \textbf{100.00} & 36.76 & 37.21\\
& Llama 3 & \textbf{87.05} & 84.38 & 21.43 & 64.29 & 76.79 & 86.16 & 28.57 & 63.84 & 64.07\\
& Mixtral-8x7B & 50.45 & 69.64 & 68.30 & 62.80 & 35.27 & 46.43 & 90.62 & 57.44 & 60.12 \\
& Claude 3 Haiku & 78.12 & 80.80 & 57.59 & \textbf{72.17} & 77.68 & \textbf{86.16} & 32.59 & \textbf{65.48} & \textbf{68.83} \\
\midrule
\multirow{9}{*}{France SSA} & BERT & 82.17 & \textbf{84.57} & \textbf{93.04} & 86.59 & 82.39 & 84.78 & 92.61 & 86.59 & 86.59\\
& RoBERTa & \textbf{85.22} & 84.13 & 90.87 & \textbf{86.74} & 81.52 & \textbf{86.09} & \textbf{93.04} & 86.88 & \textbf{86.81}\\
& CharRoBERTa & 84.35 & 80.43 & 91.30 & 85.36 & \textbf{83.04} & 83.04 & 91.96 & 86.01 & 85.69\\
\cmidrule{2-11}
& GPT-3.5 & 89.35 & \textbf{95.65} & 8.91 & 64.64 & 92.61 & \textbf{96.74} & 8.26 & \textbf{65.87} & 65.26\\
& Llama 2 & 1.96 & 15.22 & \textbf{91.52} & 36.23 & 32.39 & 55.43 & \textbf{71.96} & 53.26 & 44.75\\
& Llama 3 & \textbf{91.52} & 94.57 & 7.17 & 64.42 & 92.39 & 95.87 & 6.52 & 64.93 & 64.68\\
& Mixtral-8x7B & 71.96 & 88.70 & 38.04 & \textbf{66.23} & 68.26 & 83.26 & 39.35 & 63.62 & 64.93 \\
& Claude 3 Haiku & 89.13 & 93.91 & 13.70 & 65.58 & \textbf{96.75} & 94.78 & 4.57 & 65.36 & \textbf{65.47}\\
\bottomrule[1.5pt]
\end{tabular}
\caption{Experimental results for applying foundation language models and LLMs to the test sets of three balanced SSA datasets. We assessed gender prediction performance by calculating an accuracy score for each gender. Acc. represents the overall accuracy across genders. BERT, RoBERTa, and CharRoBERTa were fine-tuned using the training set of each SSA dataset. In contrast, we applied 0-shot prompting to evaluate other LLMs using the test sets.}
\label{tab2:gender-performance}
\end{table*}

% This is the table using duplicated names
\begin{table*}[!htb]
\centering
\footnotesize
\begin{tabular}{cccccc|cccc}
\toprule[1.5pt]
 % & & \multirow{3}{*}{First Name} & Micro F1 & Acc. & Macro F1 & Micro F1 & Acc. \\
% \multicolumn{2}{c{>{\centering\arraybackslash}p{2cm}}} & First & First+Country\\
\multicolumn{1}{c}{}&\multicolumn{1}{c}{}&\multicolumn{4}{c}{First Name}&\multicolumn{4}{c}{First Name + Year}\\
\midrule
Datasets & Models & Male & Female & Neutral & Acc. & Male & Female & Neutral & Acc. \\
\midrule
\multirow{10}{*}{US SSA} & GPT-3.5 (0-shot) & 86.30 & 92.39 & 31.80 & 55.61 & 95.21 & \textbf{93.66} & 3.92 & 41.94 \\
& Llama 2 (0-shot) & 14.94 & 33.60 & \textbf{94.23} & \textbf{63.94} & 47.80 & 62.12 & \textbf{66.70} & \textbf{61.04} \\
& Llama 3 (0-shot) & \textbf{92.53} & \textbf{93.19} & 11.89 & 45.80 & 96.26 & 93.50 & 2.02 & 41.09 \\
& Mixtral-8x7B (0-shot) & 80.84 & 91.28 & 32.06 & 54.15 & 70.59 & 92.23 & 32.49 & 51.88 \\
& Claude 3 Haiku (0-shot) & 88.89 & 91.60 & 25.85 & 52.70 & \textbf{96.74} & 90.97 & 10.60 & 45.80 \\
\cmidrule{2-10}
& GPT-3.5 (5-shot) & 84.96 & 91.92 & 43.64 & 62.06 & 65.33 & 67.35 & 4.05 & 30.06 \\
& Llama 2 (5-shot) & 24.71 & 50.40 & \textbf{86.17} & \textbf{64.46} & 36.88 & 64.98 & \textbf{68.94} & \textbf{59.93} \\
& Llama 3 (5-shot) & \textbf{92.82 }& 94.45 & 13.96 & 47.27 & \textbf{93.77} & \textbf{95.72} & 11.33 & 46.20 \\
& Mixtral-8x7B (5-shot) & 79.79 & \textbf{95.09} & 16.76 & 45.60 & 74.81 & 90.65 & 39.55 & 56.83 \\
& Claude 3 Haiku (5-shot) & 87.45 & 84.63 & 39.34 & 59.06 & 91.38 & 88.75 & 32.36 & 56.68 \\
\midrule
\multirow{10}{*}{Canada SSA} & GPT-3.5 (0-shot) & 86.36 & 78.07 & 49.08 & 54.74 & \textbf{97.27} & 78.95 & 19.00 & 30.81 \\
& Llama 2 (0-shot) & 21.82 & 28.07 & \textbf{98.62} & \textbf{86.01} & 4.55 & 8.77 & \textbf{99.82} & \textbf{83.87} \\
& Llama 3 (0-shot) & \textbf{92.73 }& 78.07 & 22.32 & 33.10 & 87.27 & \textbf{84.21} & 13.93 & 26.22 \\
& Mixtral-8x7B (0-shot) & 67.27 & \textbf{78.95} & 46.31 & 50.92 & 50.00 & 79.82 & 60.70 & 61.47 \\
& Claude 3 Haiku (0-shot) & 88.18 & \textbf{78.95} & 41.88 & 49.01 & 89.09 & 77.19 & 43.36 & 50.15 \\
\cmidrule{2-10}
& GPT-3.5 (5-shot) & 84.55 & 74.56 & 56.00 & 60.02 & \textbf{97.27} & 80.70 & 18.82 & 30.81 \\
& Llama 2 (5-shot) & 22.73 & 24.56 & \textbf{97.42} & \textbf{84.79} & 32.73 & 23.68 & \textbf{87.27} & \textbf{77.14} \\
& Llama 3 (5-shot) & \textbf{91.82} & \textbf{79.82} & 36.62 & 45.03 & 82.73 & \textbf{85.96} & 32.01 & 40.98 \\
& Mixtral-8x7B (5-shot) & 68.18 & 77.19 & 49.17 & 53.21 & 68.18 & 74.56 & 58.30 & 60.55 \\
& Claude 3 Haiku (5-shot) & 83.64 & 64.91 & 55.26 & 58.49 & 90.91 & 60.53 & 41.97 & 47.71 \\
\midrule
\multirow{10}{*}{France SSA} & GPT-3.5 (0-shot) & 78.43 & \textbf{98.31} & 16.52 & 34.30 & \textbf{90.20} & \textbf{98.31} & 3.54 & 25.84 \\
& Llama 2 (0-shot) & 3.92 & 35.59 & \textbf{89.38} & \textbf{72.61} & 27.45 & 79.66 & \textbf{74.93} & \textbf{70.16} \\
& Llama 3 (0-shot) & 74.51 & \textbf{98.31} & 4.13 & 24.50 & 90.20 & \textbf{98.31} & 0.00 & 23.16 \\
& Mixtral-8x7B (0-shot) & \textbf{82.35} & 94.92 & 14.75 & 32.96 & 88.24 & 94.92 & 28.91 & 44.32 \\
& Claude 3 Haiku (0-shot) & 78.43 & 94.92 & 10.62 & 29.40 & 88.24 & 94.92 & 6.78 & 27.62 \\
\cmidrule{2-10}
& GPT-3.5 (5-shot) & 78.43 & 98.31 & 20.35 & 37.19 & \textbf{98.04} & \textbf{100.00} & 5.01 & 28.06 \\
& Llama 2 (5-shot) & 3.92 & 33.90 & \textbf{88.20} & \textbf{71.49} & 13.73 & 47.46 & \textbf{91.15} & \textbf{76.61} \\
& Llama 3 (5-shot) & 82.35 & 98.31 & 9.44 & 29.40 & 90.20 & \textbf{100.00} & 5.01 & 27.17 \\
& Mixtral-8x7B (5-shot) & \textbf{88.24} & \textbf{100.00} & 13.57 & 33.41 & 88.24 & 94.92 & 28.91 & 44.32 \\
& Claude 3 Haiku (5-shot) & 74.51 & 86.44 & 41.00 & 50.78 & 94.12 & 96.61 & 26.25 & 43.21 \\
\bottomrule[1.5pt]
\end{tabular}
\caption{Gender prediction results of LLMs using dynamic gender label datasets under 0- and 5-shot settings. We report the gender prediction performance using accuracy for each gender. Acc. denotes the overall accuracy across genders. Appendix~\ref{prompt-template} and~\ref{robustness-evaluation} provide the prompt templates and prompt robustness evaluation for LLMs.}
\label{tab:duplicated-performance}
\end{table*}

\subsection{Gender Prediction Models}
We compared several pre-trained foundation language models with a classification head to predict the gender of first names as a multi-class classification task. Additionally, we conducted LLM-based 0-shot and 5-shot experiments to evaluate the performance of LLMs as gender classifiers. 

\textbf{Foundation Language Models.} We fine-tuned three widely used foundation language models, i.e., BERT, RoBERTa, and CharBERT, as baselines for name-based gender prediction under the same experimental settings to conduct gender prediction. Model tuning hyper-parameters are detailed in Appendix~\ref{sec:appendix}. 

\textbf{Large Language Models.} We aimed to identify the potential gender bias of LLMs in predicting gender labels given first names (plus birth year). We used five widely used LLMs for experimentation: GPT-3.5\footnote{\url{https://platform.openai.com/docs/models/gpt-3-5-turbo}} \cite{gpt3.5}, Llama 2\footnote{\url{https://llama.meta.com/llama2/}} \cite{touvron2023llama}, Llama 3\footnote{\url{https://llama.meta.com/llama3/}} \cite{llama3modelcard}, Mixtral-8x7B\footnote{\url{https://mistral.ai/news/mixtral-of-experts/}} \cite{jiang2024mixtral}, and Claude 3 Haiku\footnote{\url{https://www.anthropic.com/news/claude-3-haiku}} \cite{anthropic2024claude}. For more information about these models and the settings we used see Appendix~\ref{sec:appendix} and Appendix~\ref{llm-information}, respectively.  
% In addition to comparing LLMs and fine-tuned PLMs in general gender prediction, we perform few-shot prompting with names that have different gender labels over time using the three name gender datasets. This evaluates the effectiveness of LLMs in using birth year information for gender inference. 

\begin{figure}[!h]
  \centering
  \includegraphics[width=\columnwidth]{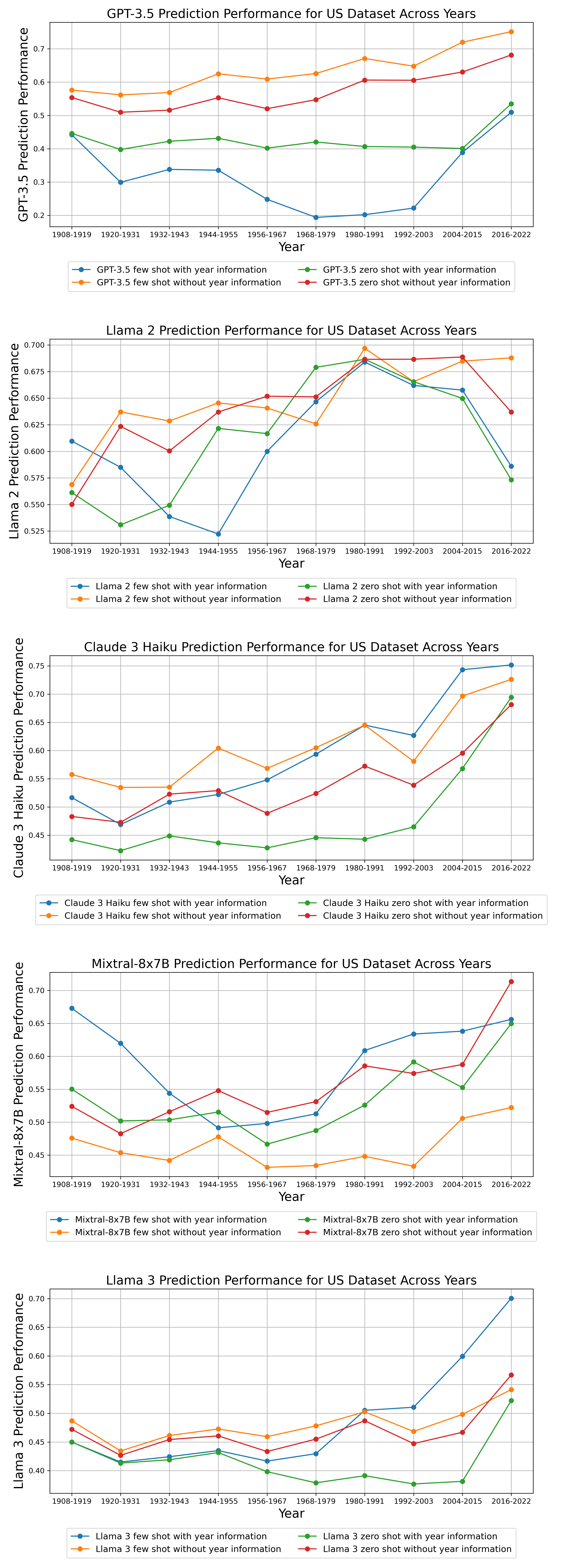}
  \caption{Temporal-level comparison of 5 LLMs using the US SSA dynamic gender label dataset given the results of Table~\ref{tab:duplicated-performance}. We report the overall accuracy of gender prediction for each year.}
  \label{duplicated-graph}
\end{figure}

\subsection{Results} \label{sec:results}
% Providing experimental results of gender prediction. We can also plot how LLMs predict the gender over years. For example, given the x-axis as the year and the y-axis as the distribution of each gender label being predicted (as we sample the SSA data with balanced gender distribution each year). % Also, compare the LLM's performance given country-level datasets. What's the difference in gender prediction accuracy in terms of country-wise names? 

% In the experimental results presented in Table~\ref{tab2:gender-performance}, we compare the gender prediction performance of fine-tuned MLMs such as BERT, RoBERTa, and CharRoBERTa (using their cased versions) with LLMs like GPT-3.5, Llama 2, Mixtral, and Claude 3 under a zero-shot setting. The models receive either the first name alone or the first name plus the birth year as input. 

\textbf{RQ1: How does the performance of LLMs versus fine-tuned foundation language models compare in first-name gender prediction?} 
Fine-tuned foundational language models predicted gender-neutral first names more accurately than LLMs under 0-shot prompting across all three datasets. As shown in Table~\ref{tab2:gender-performance}, out of all models, BERT results in the highest average accuracy for the US and Canada dataset, while RoBERTa outperformed BERT on the France dataset. Claude 3 Haiku achieved the highest accuracy among the LLMs with 0-shot prompting on all three datasets. The Llama 2 model did best on identifying gender-neutral names (100\% accuracy for Canada SSA, 99.66\% for US SSA, and 91.52\% for France SSA when using only first names as input). Llama 3 demonstrated a more balanced distribution of prediction performance across different gender categories, similar to other LLMs such as GPT-3.5, Mixtral-8x7B, and Claude 3 Haiku. However, most LLMs failed to predict gender-neutral first names in the France SSA dataset compared to the English-based datasets, with accuracies of 7.17\% for Llama 3, 8.91\% for GPT-3.5, and 13.7\% for Claude 3 Haiku. To assess the performance of gender prediction in dynamic gender label datasets (see Table~\ref{tab:duplicated-performance}), we evaluated LLMs in 0-shot and 5-shot settings, using only first names as input. Most LLMs showed higher accuracy in gender prediction when provided with 5 labeled name-gender pairs through in-context learning compared to the 0-shot setting across all datasets. 

% Regarding the gender label prediction performance for dynamic gender label datasets, 

\textbf{RQ2: How does adding the birth year impact gender prediction accuracy?} The effectiveness of the input variation (i.e., first name + birth year) varied among different language models. Incorporating birth years as  an additional input feature improved the prediction accuracy of foundational language models compared to the first-name-only setting (Table~\ref{tab2:gender-performance}). However, most LLMs showed a decline in accuracy when birth years were added, particularly in predicting gender-neutral names. Despite this trend, Mixtral-8x7B consistently improved its prediction accuracy for gender-neutral names across all three datasets by adding birth year information. Similarly, the overall accuracy of Llama 2 increased, with improvements of 12\% and 17\% in the US and France SSA datasets, respectively.
% Notably, Llama 3 demonstrated a significant decrease in gender-neutral name predictions compared with Llama 2 but achieved superior performance of binary genders, especially in male name predictions without temporal information. 
% Most of the overall accuracy of Llama 3, Mixtral-8x7B, and Claude 3 Haiku decreased across the three datasets compared to first name only scenario.

% The accuracy scores of Llama 2 under US SSA and France SSA datasets are different. We manually checked the generated results and found some outliers in Llama 2's gender predictions. We will report these cases in Appendix [REF]. The Llama 2 generates some irrelevant content instead of gender types. Therefore, the irrelevant contents predicted by Llama 2 are out of the confusion matrix and the number of gender types is beyond 3, so the total number of each true gender label is different, which derives in not equal recall and accuracy scores. 
Additionally, including birth years decreased the accuracy of predicting gender-neutral names in both 0- and 5-shot settings across all datasets (Table~\ref{tab:duplicated-performance}), except for the Mixtral-8x7B model, which increased the gender prediction accuracy by adding birth years. The accuracy of GPT-3.5 and Llama 3 in predicting gender-neutral names dropped when adding the birth year among all three datasets.

We observed varying trends in prediction accuracy over time across 5 LLMs (Figure~\ref{duplicated-graph}). The accuracy of gender prediction using the US SSA dynamic gender label dataset has increased in recent years for most LLMs, including Llama3, Mixtral-8x7B, Claude 3 Haiku, and GPT-3.5. In particular, GPT-3.5 performed better without than with birth years, suggesting that incorporating recent birth year information in the US SSA dataset did not enhance predictive accuracy. The over-time results in Figure~\ref{duplicated-graph} indicated that most LLMs were better at predicting the genders of more recent first names. The over-time comparison of the other two datasets was provided in Appendix~\ref{temporal-trends}.

\section{Discussion} \label{sec:discussion}
% At a high level, we found:  
% Shown in Table~\ref{tab2:gender-performance}, the BERT model took the leading place in predicting gender labels in English-speaking datasets: the US and Canada. This is likely because BERT's training corpus predominantly consists of the BookCorpus and English Wikipedia, which may not extensively cover French names and their associated gender labels. However, in the case of the France SSA dataset, the RoBERTa model demonstrated superior performance compared to BERT, especially for RQ 2 results. We assume that the more varied datasets for RoBERTa model pre-training may help model distinguish , which includes CC-News, OpenWebText, and Stories, in addition to the corpora used for BERT. 

\paragraph{LLMs are poor at accurately predicting gender.} Gender bias occurs in LLMs when performing name-based gender predictions, which shows varying performance in predicting non-binary gender labels. Llama 2 categorizes nearly all names as neutral genders, with first names only as input. This tendency may result from Llama 2's training approach, which used reward modeling to promote more inclusive responses, where initial model outputs are adjusted based on human feedback to maximize inclusivenes and factual accuracy \cite{touvron2023llama}. The rewarding process allows the model to better align with modern datasets' nuanced and inclusive expectations.

\paragraph{Including temporal information mostly degrades accuracy.} When providing dynamic gender label datasets with birth year information, the gender-prediction performance of most LLMs decreased, especially for gender-neutral names. However, Mixtral-8x7B showed an increase in overall accuracy when birth years were added in 0- and 5-shot settings. We hypothesize that Mixtral-8x7B can better use temporal data as a reference for gender prediction because it is trained with more numerical information. Although Llama 2 outperformed other LLMs in predicting gender-neutral names, it exhibited biased prediction results, often classifying most names as gender-neutral. We assume Llama 2's Reinforcement Learning with Human Feedback (RLHF) approach \cite{touvron2023llama} guides the model to generate more inclusive responses. When Llama 2 is unsure about a name's gender, it may default to labeling it as neutral, potentially reducing prediction accuracy for gender-neutral names. 

\paragraph{LLMs have worst performance on gender-neutral names.} We also find that most tested LLMs have more difficulties in predicting gender-neutral first names than binary genders, which may stem from the training data of LLMs that primarily includes binary gender labels in the training documents \cite{touvron2023llama}. Llama 3, in particular, performed poorly overall across all three datasets with different input variations (i.e., first names with or without birth years). As detailed in Appendix~\ref{dataset-creation}, the datasets used for dynamically labeling genders were imbalanced, with gender-neutral names being the majority. Specifically, the total numbers of gendered names for the US, Canada, and France SSA datasets were 3,996, 1,308, and 449, respectively, with around 58.1\%, 82.9\%, and 75.5\% being gender-neutral. Consequently, Llama 3 underperformed in overall prediction accuracy compared to other LLMs due to its poor accuracy in predicting neutral genders despite performing better in predicting binary genders.

\paragraph{LLM performance is biased towards recent year patterns.} Based on the over-time comparison of the US SSA dataset (Figure~\ref{duplicated-graph}), we hypothesize that the improved prediction performance of LLMs for recent data can be attributed to the increased volume of training data from recent years. We assume that the training data of LLMs is unbalanced, predominantly consisting of recent data, potentially explaining the higher gender prediction accuracy of LLMs in recent years. The comparison of balanced SSA datasets and dynamic gender label datasets shown in Table~\ref{tab2:gender-performance} and Table~\ref{tab:duplicated-performance} indicates that LLMs face challenges not only with predicting gender-neutral names but also with dynamically changing gender associations for the same names. This issue likely originates from the inherent limitations of the pre-training approach and data used in LLMs. These models tend to memorize training data, which lacks inferential capability, rather than adapting well to names with evolving gender labels over time. Overall, most LLMs better predict female names than male names, and the accuracy of gender prediction is higher for English-based first names in the US and Canada SSA datasets than in the France SSA.

\paragraph{Suggestions for practitioners} As we have highlighted in this work, LLMs have a biased and in-accurate understanding of names and hence we should be careful about using them for gender inference related tasks, even at an aggregate level. Furthermore, when dealing with temporal and especially historical data, LLM's name-based gender understanding may be limited and hence their usage for aggregated data analysis is likely to lead to incorrect results.

% To learn more about the performance changes across different years, Figure~\ref{duplicated-graph-chatgpt} shows that it is evident that when the year of birth is included as additional reference information in the prompts, GPT-3.5's performance deteriorates compared to settings without this information, particularly in the US SSA and Canada datasets. However, an improvement in prediction accuracy is observed in the France dataset with more recent years, as well as in the US SSA dataset. 

\section{Conclusion}
% To assess the gender inclusiveness of LLMs in the task of first-name-based gender prediction, we evaluated the performance of widely-used LLMs using three national SSA datasets. 
% To assess the gender bias of LLMs in gender prediction based on first names, we evaluated the performance of widely-used LLMs using three national SSA datasets. Our findings demonstrated that most LLMs struggled to classify gender-neutral names accurately. In particular, Llama 2 tends to classify all names as gender-neutral, likely as a precaution to avoid gender-related biases. Even when additional information, such as birth year, is provided, the models still could not effectively leverage this demographic data in their predictions. We conclude that LLMs may be unreliable for gender-intensive downstream tasks, such as named entity labeling, recommendation systems, and others, compared to traditional MLMs equipped with a classification layer. We recommend exercising caution when using LLMs for gender-related topics to avoid potential issues, particularly in scenarios involving non-binary identities. 
This study underscores the limited performance of LLMs as classifiers in predicting gender-neutral names compared to binary genders and the challenges posed by the inherent biases in the datasets used to train LLMs, which may lead to unbalanced gender prediction results. By introducing a ``neutral'' category, we have taken a step towards more inclusive gender prediction. However, our findings revealed that LLMs may struggle recognizing gender-neutral names, especially for non-English first names. Despite efforts to enhance LLMs' predictive capabilities by including temporal data, there were no meaningful improvements in gender prediction accuracy, especially for gender-neutral names. This suggests a fundamental limitation of current LLMs and training datasets when adapting to the complexities of gender identities. In future studies, we plan to expand our work by using more inclusive gender categories (e.g., cisgender and transgender) to thoroughly assess gender bias in LLMs across various NLP downstream tasks, including sentiment analysis and coreference resolution.

\section{Bias Statement}
Our study investigates gender bias in LLMs and fine-tuned foundation language models when predicting the gender of names by introducing a ``neutral'' category alongside the traditional binary classification of male and female gender labels. Traditionally, the binary gender classification system has not accounted for gender-neutral names. This exclusion arises from imbalanced training data and fixed representations of gender (i.e., female and male), causing LLMs to be prone to classify names into binary gender labels.

When using LLMs in name-based gender prediction tasks, they generally consider only two gender labels, thereby restricting the scope of gender-related analysis. This binary approach perpetuates potential biases in areas associated with fixed gender representations \cite{ liu2023gender, teich2022citation}, e.g., how male and female authors express sentiment \cite{jentzsch2022gender} or how male and female researchers face different challenges in academia \cite{vanhelene2024inferring}. However, this binary labeling of gender overlooks individuals with gender-neutral names, which could encompass both female and male identities, thereby missing valuable insights from a more inclusive perspective. Our work considers more inclusive gender labeling by examining the accuracy of gender-neutral name predictions using LLMs while also providing insights into factors that may lead to biased gender prediction results (i.e., poorer prediction for neutral names compared to binary names) in these models.

\section*{Limitations}
Our study's limitations are as follows:
(1) Our assessment was limited to specific countries, i.e., the US, Canada, and France, not considering a broad spectrum of countries and cultures, particularly in Asia and Africa. This limitation may affect the generalizability of our findings across different cultural and linguistic contexts. 
(2) The dataset preparation involved a subjective threshold to determine gender-neutral names, defined as names where the gender frequency for both males and females is greater than 10\%. This choice may impact the reliability and consistency of the presented findings. 
(3) The prompt templates employed for interacting with LLMs were not optimized, which may lead to variations in results with different prompt formulations. This indicates a potential variability in LLMs' performance that could impact the robustness of our conclusions, as LLMs are sensitive to prompt design.

% \section*{Acknowledgements}

% Bibliography entries for the entire Anthology, followed by custom entries
%\bibliography{anthology,custom}
% Custom bibliography entries only
% \bibliography{custom}

% \clearpage
\appendix

\begin{table}[h!]
% \centering
\footnotesize
\scalebox{0.74}{
\begin{tabular}{ccccccc}
% \hline
\toprule[1.5pt]
\textbf{Datasets} & \textbf{\# Names} & \textbf{Year span} & \textbf{Train} & \textbf{Val} & \textbf{Test} & \textbf{Overall}\\
\midrule
US SSA & 300 & 1914 - 2022 & 78480 & 9810 & 9810 & 98100 \\
\midrule
Canada SSA & 273 & 2013 - 2020 & 5232 & 648 & 672 & 6552\\
\midrule
France SSA & 32 & 1908 - 2022 & 8625 & 1035 & 1380 & 11040\\
\bottomrule[1.5pt]
\end{tabular}}
\caption{\label{dataset-statistics}
Statistics of balanced SSA datasets. \# Names represent the number of names per gender per year.
}
\end{table}

\begin{table}[h!]
\centering
\footnotesize
\begin{tabular}{cccc}
% \hline
\toprule[1.5pt]
\textbf{Datasets} & \textbf{\# Neutral} & \textbf{\# Male} & \textbf{\# Female}\\
\midrule
US SSA & 2321 & 1044 & 631 \\
\midrule
Canada SSA & 1084 & 110 & 114\\
\midrule
France SSA & 339 & 59 & 51\\
\bottomrule[1.5pt]
\end{tabular}
\caption{\label{duplicated-data}
Statistics of dynamic gender label datasets.
}
\end{table}

\section{Dataset Statistics}
\label{dataset-creation}
Overall training and testing dataset statistics were reported in Table~\ref{dataset-statistics}. We split the train/val/test sets into 80\%/10\%/10\% of the data. We found that gender-neutral names have increased in both the US and Canada SSA datasets over time and surged in more recent years (i.e., after 2000). 

Dataset statistics of dynamic gender labels extracted from the three datasets' test sets are reported in Table~\ref{duplicated-data}. Note that the Canada SSA dataset only contained 63 first names whose gender labels changed over time in the test set and 50 in the validation set, which was insufficient for evaluating LLMs' performance in dynamic gender prediction. Therefore, we used the training set to extract the names with dynamic gender labels for the Canada SSA dataset.

\section{Experimental Settings}
\label{sec:appendix}

In foundation language model fine-tuning, we set the maximum length of the tokenizer to 32 across all three models since the results won't change with an increase in the maximum input length. We fine-tuned foundation language models through 7 epochs, and the batch size for either training or validation was 128. We set the warm-up ratio to 0.1 and the learning rate toas 2e-5. The foundation language models included BERT (\texttt{bert-base-cased}), RoBERTa (\texttt{roberta-base}), and CharRoBERTa. We chose the cased models because they are case-sensitive and can distinguish names such as ``huntley'' and ``Huntley''.

For the model settings of LLMs, we applied GPT-3.5  (\texttt{gpt-3.5-turbo-instruct}), Llama 2 (\texttt{meta/llama-2-70b-chat}), Llama 3 (\texttt{meta/meta-llama-3-70b-instruct}), Mixtral-8x7B (\texttt{mixtral-8x7b-instruct-v0.1}), and Claude 3 Haiku (\texttt{claude-3-haiku-20240307}) for name gender prediction tasks.

\begin{table*}[!h]
\centering
\footnotesize
\scalebox{0.90}{
\begin{tabularx}{500pt}{lXX}
\toprule
\textbf{Experimental Setting} & \textbf{RQ 1} & \textbf{RQ 2} \\
\midrule
0-shot & Predict the gender association of the given name. \texttt{\textbackslash n}Use the following labels for classification: \texttt{\textbackslash n}Male: The name is predominantly associated with males. \texttt{\textbackslash n}Female: The name is predominantly associated with females. \texttt{\textbackslash n}Neutral: The name is not predominantly associated with any single gender and is considered neutral. \texttt{\textbackslash n}Your outputs should be all in lowercase and can only output gender from male, female, or neutral. \texttt{\textbackslash n}Name: + \texttt{\{name\}} + \texttt{\textbackslash n}Gender: & Predict the gender association of the given name, considering the year of birth as an additional reference. \texttt{\textbackslash n}The provided names appear more than once across different years of birth as they may be labeled in different genders given the change in the predominant gender of names. \texttt{\textbackslash n}Use the following labels for classification: \texttt{\textbackslash n}Male: The name is predominantly associated with males. \texttt{\textbackslash n}Female: The name is predominantly associated with females. \texttt{\textbackslash n}Neutral: The name is not predominantly associated with any single gender and is considered neutral. \texttt{\textbackslash n}Your outputs should be all in lowercase and can only output gender from male, female, or neutral. \texttt{\textbackslash n}Name: + \texttt{\{name\}} + \texttt{\textbackslash n}Year of Birth: + \texttt{\{year\}} + \texttt{\textbackslash n}Gender:\\
\midrule
5-shot (US SSA) & Predict the gender association of the given name. \texttt{\textbackslash n}The provided names appear more than once. \texttt{\textbackslash n}Use the following labels for classification: \texttt{\textbackslash n}Male: The name is predominantly associated with males. \texttt{\textbackslash n}Female: The name is predominantly associated with females. \texttt{\textbackslash n}Neutral: The name is not predominantly associated with any single gender and is considered neutral. \texttt{\textbackslash n}Please note that first names can be labeled in different genders over time. \texttt{\textbackslash n}Here are five pairs of examples of first names and genders: \texttt{\textbackslash n}Pair 1: Name: Christie, Gender: Neutral; Name: Christie, Gender: Female
Pair 2: Name: Jan, Gender: Neutral; Name: Jan, Gender: Male
Pair 3: Name: Bee, Gender: Female; Name: Bee, Gender: Neutral
Pair 4: Name: Kasen, Gender: Neutral; Name: Kasen, Gender: Male
Pair 5: Name: Mel, Gender: Male; Name: Mel, Gender: Neutral
\texttt{\textbackslash n}Your outputs should be all in lowercase and can only output gender from male, female, or neutral. \texttt{\textbackslash n}Name:  + \texttt{\{name\}} + \texttt{\textbackslash n}Gender: & Predict the gender association of the given name, considering the year of birth as an additional reference. \texttt{\textbackslash n}The provided names appear more than once. \texttt{\textbackslash n}Use the following labels for classification: \texttt{\textbackslash n}Male: The name is predominantly associated with males. \texttt{\textbackslash n}Female: The name is predominantly associated with females. \texttt{\textbackslash n}Neutral: The name is not predominantly associated with any single gender and is considered neutral. \texttt{\textbackslash n}Please note that first names can be labeled in different genders over time. \texttt{\textbackslash n}Here are five pairs of examples of first names and genders: \texttt{\textbackslash n}Pair 1: Name: Christie, Year of Birth: 1919, Gender: Neutral; Name: Christie, Year of Birth: 1949, Gender: Female
Pair 2: Name: Jan, Year of Birth: 1966, Gender: Neutral; Name: Jan, Year of Birth: 2012, Gender: Male
Pair 3: Name: Bee, Year of Birth: 1952, Gender: Female; Name: Bee, Year of Birth: 1989, Gender: Neutral
Pair 4: Name: Kasen, Year of Birth: 2000, Gender: Neutral; Name: Kasen, Year of Birth: 2006, Gender: Male
Pair 5: Name: Mel, Year of Birth: 1947, Gender: Male; Name: Mel, Year of Birth: 2007, Gender: Neutral
\texttt{\textbackslash n}Your outputs should be all in lowercase and can only output gender from male, female, or neutral. \texttt{\textbackslash n}Name: + \texttt{\{name\}} + \texttt{\textbackslash n}Year of Birth: + \texttt{\{year\}} + \texttt{\textbackslash n}Gender:\\
5-shot (Canada SSA) & ...Pair 1: Name: Nyjah, Gender: Neutral; Name: Nyjah, Gender: Male
Pair 2: Name: Kendell, Gender: Neutral; Name: Kendell, Gender: Male
Pair 3: Name: Arshia, Gender: Neutral; Name: Arshia, Gender: Male
Pair 4: Name: Lennix, Gender: Neutral; Name: Lennix, Gender: Female
Pair 5: Name: Kirat, Gender: Male; Name: Kirat, Gender: Neutral...
& ...Pair 1: Name: Nyjah, Year of Birth: 2014, Gender: Neutral; Name: Nyjah, Year of Birth: 2016, Gender: Male
Pair 2: Name: Kendell, Year of Birth: 2014, Gender: Neutral; Name: Kendell, Year of Birth: 2016, Gender: Male
Pair 3: Name: Arshia, Year of Birth: 2014, Gender: Neutral; Name: Arshia, Year of Birth: 2018, Gender: Male
Pair 4: Name: Lennix, Year of Birth: 2013, Gender: Neutral; Name: Lennix, Year of Birth: 2018, Gender: Female
Pair 5: Name: Kirat, Year of Birth: 2013, Gender: Male; Name: Kirat, Year of Birth: 2014, Gender: Neutral...\\
5-shot (France SSA) & ...Pair 1: Name: CARMEL, Gender: Male; Name: CARMEL, Gender: Neutral 
Pair 2: Name: LIE, Gender: Male; Name: LIE, Gender: Neutral Pair 3: Name: JESSY, Gender: Female; Name: JESSY, Gender: Neutral
Pair 4: Name: ANH, Gender: Neutral; Name: ANH, Gender: Male
Pair 5: Name: FIDELE, Gender: Neutral; Name: FIDELE, Gender: Female... & ...Pair 1: Name: CARMEL, Year of Birth: 1920, Gender: Male; Name: CARMEL, Year of Birth: 1951, Gender: Neutral
Pair 2: Name: LIE, Year of Birth: 1922, Gender: Male; Name: LIE, Year of Birth: 1931, Gender: Neutral
Pair 3: Name: JESSY, Year of Birth: 1960, Gender: Female; Name: JESSY, Year of Birth: 1975, Gender: Neutral
Pair 4: Name: ANH, Year of Birth: 1995, Gender: Neutral; Name: ANH, Year of Birth: 2006, Gender: Male
Pair 5: Name: FIDELE, Year of Birth: 1918, Gender: Neutral; Name: FIDELE, Year of Birth: 1945, Gender: Female...\\
\bottomrule
\end{tabularx}}
\caption{Task-oriented prompt templates of LLMs in 0-shot and 5-shot settings for RQ 1 (w/o birth year) and RQ 2 (w/ birth year). For clarity, we report only the 5-shot example pairs for Canada and France's SSA datasets, as the prompt templates are the same as those used for the 5-shot US SSA dataset.}
\label{tab:prompt-templates}
\end{table*}

\section{LLMs for Gender Prediction} \label{llm-information}
We applied the 5 LLMs for name-based gender prediction using three country-level SSA datasets.

\textbf{GPT-3.5.} GPT-3.5 is an autoregressive generation model developed by OpenAI \cite{gpt3.5}. The model (\texttt{gpt-3.5-turbo-instruct}) has been tuned through an instruction-tuning technique and aims to generate human-preferred responses.

\textbf{Llama 2.} Llama 2 is a collection of open-source chat models developed by Meta, ranging from 7 to 70B parameters \cite{touvron2023llama}. It was trained on 2 trillion tokens of publicly available data and tuned through over one million new human-annotated examples. We applied \texttt{llama-2-chat} for our experiments.

\textbf{Llama 3.} Following Llama 2, Llama 3 is a series of pre-trained and instruction-tuned autoregressive models in 8 and 70B sizes \cite{llama3modelcard}. The training data of Llama 3 is over seven times larger than Llama 2, reaching over 15 trillion tokens of data and over 10M human-annotated examples. 

\textbf{Mixtral-8x7B.} Mixtral-8x7B is a pre-trained generative Sparse Mixture of Experts \cite{jiang2024mixtral}. The Mixtral-8x7B outperformed Llama 2 70B on most benchmarks and can handle English, French, Italian, German, and Spanish, which is helpful when predicting French name genders.

\textbf{Claude 3 Haiku.} Claude 3 family is a series of close-source language models, including three state-of-the-art models in ascending order of capability: Claude 3 Haiku, Claude 3 Sonnet, and Claude 3 Opus \cite{anthropic2024claude}. Claude 3 Haiku is the fastest, most compact model for near-instant responsiveness. We used Claude 3.

\section{Prompt Templates for LLMs} \label{prompt-template}
We reported the prompt templates for the experiments of LLMs in 0- and 5-shot settings for RQ 1 and RQ 2 in Table~\ref{tab:prompt-templates}. For RQ 1, we used ``First Name'' for gender prediction. For RQ 2, we provided ``First Name'' and ``Year of Birth'' as input.

% \begin{tcolorbox}[enhanced,boxrule=1pt,colback=white,colframe=black,boxsep=5pt,arc=4pt]
% \noindent \textbf{Zero-shot Prompt Template for RQ 1}\\

% \noindent Predict the gender association of the given name. Use the following labels for classification:\\

% Male: The name is predominantly associated with males.\\
% Female: The name is predominantly associated with females.\\
% Neutral: The name is not predominantly associated with any single gender and is considered neutral.\\

% Your outputs should be all in lowercase and can only output gender from male, female, or neutral. \\
% "\textbackslash n Name: " + name + "\textbackslash n Gender: "\\

% \noindent \textbf{Zero-shot Prompt Template for RQ 2}\\

% \noindent Predict the gender association of the given name, considering the year of birth as an additional reference. \\

% The provided names appear more than once across different years of birth as they may be labeled in different genders given the change in the predominant gender of names.\\
% Use the following labels for classification:\\ \\
% Male: The name is predominantly associated with males.\\
% Female: The name is predominantly associated with females.\\
% Neutral: The name is not predominantly associated with any single gender and is considered neutral.\\

% Your outputs should be all in lowercase and can only output gender from male, female, or neutral. \\
% "\textbackslash n Name: " + name + "\textbackslash n Year of Birth: " + year + "\textbackslash n Gender: "
% \end{tcolorbox}

In the 5-shot setting, we randomly chose five name-gender pairs from the three SSA datasets, using the number 42 as the random seed. We selected names that appeared at least twice and were assigned different genders in different years.

\section{Prompt Robustness Evaluation} \label{robustness-evaluation}
The effectiveness of prompts designed for LLM-based experiments is crucial for the performance of downstream natural language processing tasks, as highlighted by \citet{zhou2022large, zhu2023promptbench}. Therefore, we developed two prompt templates inspired by \citet{zhu2023promptbench}: task-oriented and role-oriented prompts, to evaluate the robustness of LLM gender prediction performance. The task-oriented prompt was the same as introduced in Appendix~\ref{prompt-template}. 

\begin{tcolorbox}[enhanced,boxrule=1pt,colback=white,colframe=black,boxsep=5pt,arc=4pt]
\noindent \textbf{0-shot Role-Based Prompt for RQ 1}\\
\noindent In the role of a first name gender prediction tool, classify names based on their gender association using the following gender labels:\\

Male: The name is predominantly associated with males.\\
Female: The name is predominantly associated with females.\\
Neutral: The name is not predominantly associated with any single gender and is considered neutral.\\

The provided names appear more than once. Your outputs should be all in lowercase and can only output gender from male, female, or neutral. 
"\textbackslash n Name: " + name + "\textbackslash n Gender: "
\end{tcolorbox}

\begin{tcolorbox}[enhanced,boxrule=1pt,colback=white,colframe=black,boxsep=5pt,arc=4pt]
\noindent \textbf{0-shot Role-Based Prompt for RQ 2}

\noindent In the role of a first name gender prediction tool, classify names based on their gender association using the following gender labels:\\

Male: The name is predominantly associated with males.\\
Female: The name is predominantly associated with females.\\
Neutral: The name is not predominantly associated with any single gender and is considered neutral.\\

Consider the year of birth as an additional reference. The provided names appear more than once across different years of birth as they may be labeled in different genders given the change in the predominant gender of names. \\

Your outputs should be all in lowercase and can only output gender from male, female, or neutral. 
"\textbackslash n Name: " + name + "\textbackslash n Year of Birth: " + year + "\textbackslash n Gender: "
\end{tcolorbox}

Above are examples of role-based prompts used in RQ 1 and 2 under the 0-shot setting. The 5-shot examples are the same as we applied in task-oriented prompts. We provided first names after ``Name'' and guided LLMs to output genders after ``Gender''. 

We evaluated the robustness of prompts using GPT-3.5 on the France SSA dynamic gender label dataset referenced in Table~\ref{tab:duplicated-performance}. As shown in Table~\ref{tab:prompt-robustness}, our results indicate that in the 0-shot setting, both prompts exhibited similar performance for predicting male and female genders. However, using the task-oriented prompt showed a better performance in predicting gender-neutral names than using the role-oriented prompt. Given that over 75\% of names in the French dataset were gender-neutral, even minor discrepancies in the ``Neutral'' category can significantly impact the overall accuracy. While the role-oriented prompt yielded better predictions for binary gender predictions when only the first names were provided, its overall accuracy still fell behind the task-oriented setting in both experimental setups. Notably, incorporating birth year as an additional feature for name gender prediction reduced the differences between various prompt templates, particularly for the performance of gender-neutral names (Table~\ref{tab:prompt-robustness}). 

We also assessed the impact of including ``Country'' information in the gender prediction prompt using the France dataset. The results indicated no significant difference (i.e., the variation in overall accuracy is within 2\%) when incorporating the original country of the given names in both 0-shot and 5-shot settings. 

\begin{table*}[!h]
\centering
\footnotesize
\begin{tabular}{ccccc|cccc} 
\toprule[1.5pt]
 % & & \multirow{3}{*}{First Name} & Micro F1 & Acc. & Macro F1 & Micro F1 & Acc. \\
% \multicolumn{2}{c{>{\centering\arraybackslash}p{2cm}}} & First & First+Country\\
\multicolumn{1}{c}{}&\multicolumn{4}{c}{First Name}&\multicolumn{4}{c}{First Name + Year}\\
\midrule
Models & Male & Female & Neutral & Acc. & Male & Female & Neutral & Acc.\\
\midrule
Task-o Oriented Prompt (0-shot) & 78.43 & 98.31 & 16.52 & 34.30 & 90.20 & 98.31 & 3.54 & 25.84\\
Role-o Oriented Prompt (0-shot) & 78.43 & 98.31 &9.73 & 29.18 & 88.24 & 98.31 & 3.54 & 25.61\\
\midrule
Task-o Oriented Prompt (5-shot) & 78.43 & 98.31 & 20.35 & 37.19 & 98.04 & 100.00 & 5.01 & 28.06\\
Role-o Oriented Prompt (5-shot) & 90.20 & 100.00 & 17.11 & 36.30 & 92.16 & 100.00 & 4.42 & 26.95\\
\bottomrule[1.5pt]
\end{tabular}
\caption{Prompt robustness evaluation of name gender prediction using GPT-3.5 under the France dynamic gender label dataset.}
\label{tab:prompt-robustness}
\end{table*}

%%% CANADA 
\begin{figure}
  \centering
  \includegraphics[width=\columnwidth]{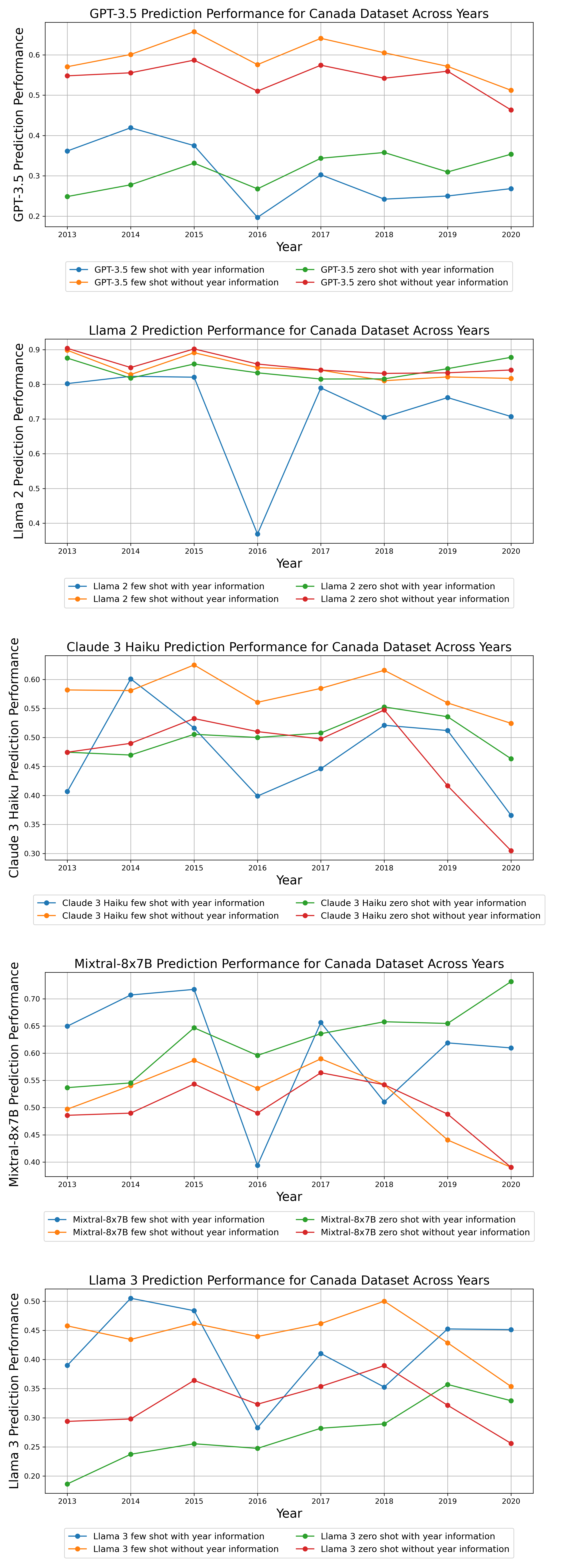}
  \caption{Temporal-level comparison of all LLMs across Canada SSA dynamic gender label dataset given the results of Table~\ref{tab:duplicated-performance}.}
  \label{duplicated-graph-canada}
\end{figure}

%%% France
\begin{figure}
  \centering
  \includegraphics[width=\columnwidth]{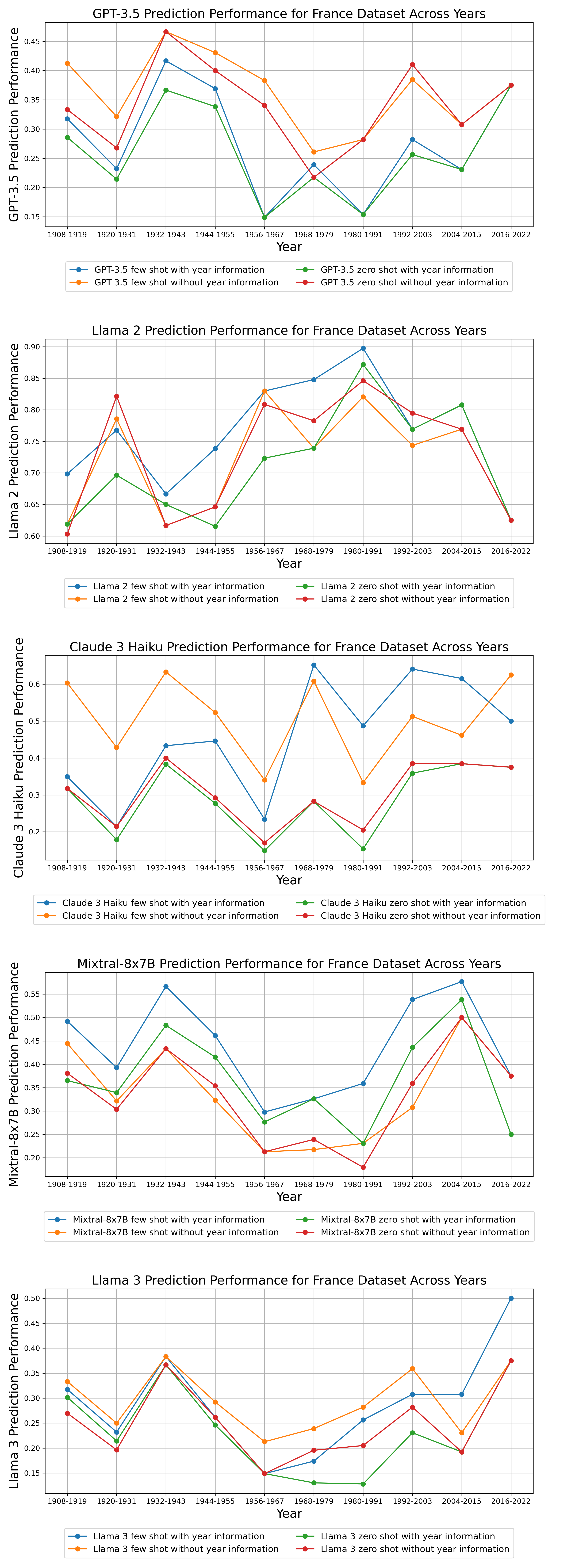}
  \caption{Temporal-level comparison of all LLMs across France SSA dynamic gender label dataset given the results of Table~\ref{tab:duplicated-performance}.}
  \label{duplicated-graph-france}
\end{figure}

\section{Over-time Trends of LLM Performances} \label{temporal-trends}
In Figures~\ref{duplicated-graph-canada} and~\ref{duplicated-graph-france}, we presented the trends in gender prediction accuracy for Canada and France using dynamic gender label datasets across five different LLMs. Generally, the performance of these LLMs varied over time for both datasets. Notably, models that did not incorporate temporal information tended to perform better, yielding more stable accuracy rates over the years than models that included birth year data. Figure~\ref{duplicated-graph-canada} also indicated that the LLMs were less effective at predicting names from more recent years. In particular, GPT-3.5 demonstrated that omitting temporal information led to higher gender prediction performance consistently over the years than including it.
% \section{Gender-Specific Performance}
% We report each model's performance in each gender type. As shown in Table[REF], there is no clear conclusion given the observation of model's performance across different gender types, especially for fine-tuning based models. However, it is worth to mention that Llama 2 tends to predict a name as 'neutral', which achieves the highest accuracy in all three datasets in both RQ 1 and RQ 2. 

\end{document}